\documentclass[conference]{IEEEtran}
\IEEEoverridecommandlockouts
\usepackage{cite}
\usepackage{amsmath,amssymb,amsfonts}
\usepackage{algorithmic}
\usepackage{graphicx}
\usepackage{textcomp}
\usepackage[hidelinks]{hyperref}
\usepackage{subcaption}
\usepackage{xcolor}
\def\BibTeX{{\rm B\kern-.05em{\sc i\kern-.025em b}\kern-.08em
    T\kern-.1667em\lower.7ex\hbox{E}\kern-.125emX}}
\begin{document}
\bstctlcite{MyBSTcontrol}

\title{PromptMix: Text-to-image diffusion models enhance the performance of lightweight networks}

\author{Arian~Bakhtiarnia,
        Qi~Zhang,
        and~Alexandros~Iosifidis
\thanks{Arian Bakhtiarnia, Qi Zhang and Alexandros Iosifidis are with DIGIT, the Department of Electrical and Computer Engineering, Aarhus University, Aarhus, Midtjylland, Denmark (e-mail: arianbakh@ece.au.dk; qz@ece.au.dk; ai@ece.au.dk).}
\thanks{This work was funded by the European Union’s Horizon 2020 research and innovation programme under grant agreement No 957337, and by the Danish Council for Independent Research under Grant No. 9131-00119B.
}}

\maketitle

\begin{abstract}
Many deep learning tasks require annotations that are too time consuming for human operators, resulting in small dataset sizes. This is especially true for dense regression problems such as crowd counting which requires the location of every person in the image to be annotated. Techniques such as data augmentation and synthetic data generation based on simulations can help in such cases. In this paper, we introduce PromptMix, a method for artificially boosting the size of existing datasets, that can be used to improve the performance of lightweight networks. First, synthetic images are generated in an end-to-end data-driven manner, where text prompts are extracted from existing datasets via an image captioning deep network, and subsequently introduced to text-to-image diffusion models. The generated images are then annotated using one or more high-performing deep networks, and mixed with the real dataset for training the lightweight network. By extensive experiments on five datasets and two tasks, we show that PromptMix can significantly increase the performance of lightweight networks by up to 26\%.
\end{abstract}


\section{Introduction}
\label{sec:intro}

Lightweight networks are essential for deploying deep learning models on IoT devices and smart cities, especially for computaionally demanding tasks such as crowd counting where the resolutions of input images are very high \cite{https://doi.org/10.1002/itl2.187}. However, lightweight networks are typically less accurate compared to their heavyweight state-of-the-art counterparts. Some methods approach this problem by proposing alternative lightweight architectures that achieve a better trade-off between performance and efficiency, such as the various versions of MobileNet \cite{DBLP:journals/corr/HowardZCKWWAA17, 8578572, 9008835}. On the other hand, the training procedure can sometimes be modified in order to enhance existing lightweight architectures.

\begin{figure*}
  \centering
   \includegraphics[width=\linewidth]{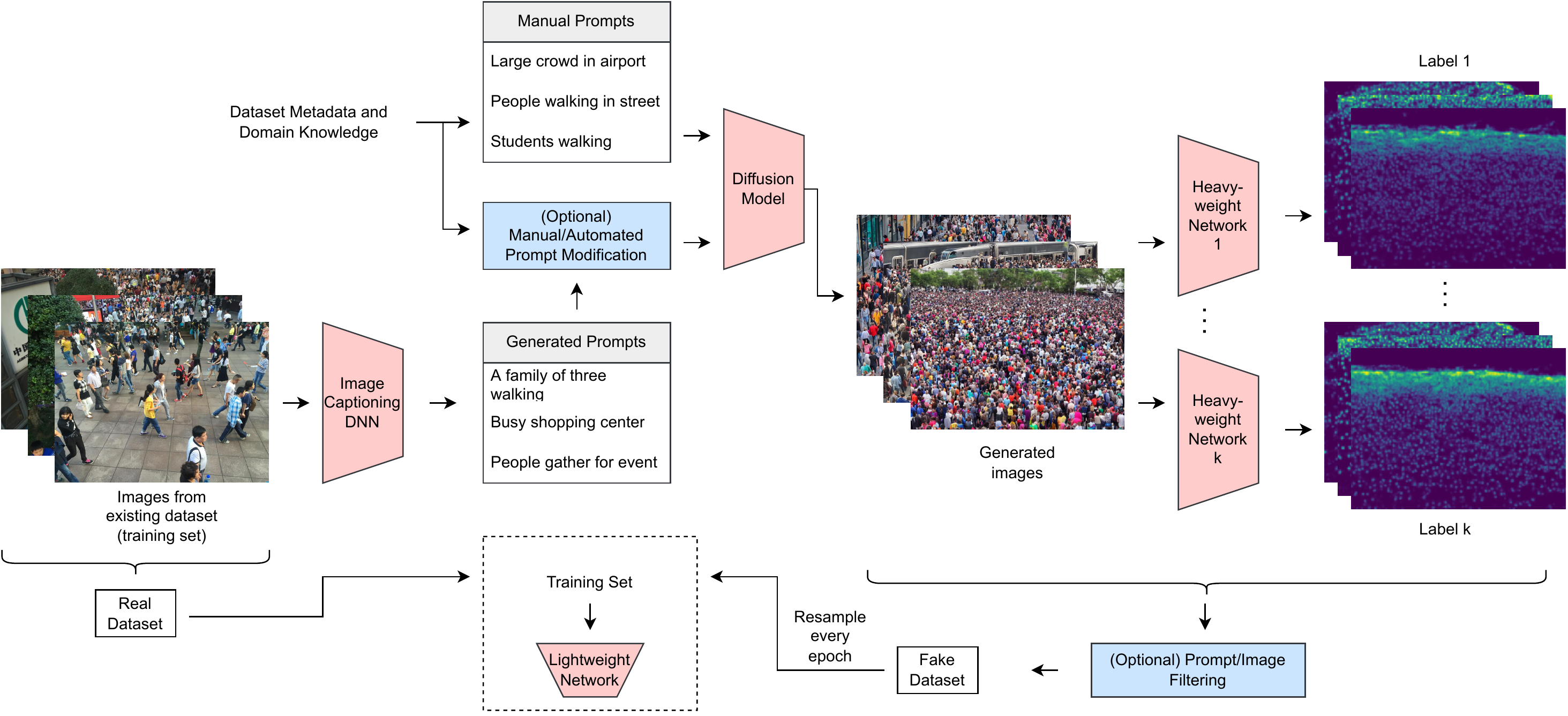}
   \caption{Overview of PromptMix.}
   \label{fig:overview}
\end{figure*}

It is well known that the amount of available data for training highly impacts the quality of deep learning models, and more data almost always leads to better performance \cite{LeCun2015}. In most cases, obtaining more data is difficult since manual labelling is a time-consuming process. Therefore, several techniques exist in the literature aiming to modify and boost the training data that is already available, such as data augmentation \cite{antoniou2017data} and the use of simulation environments \cite{Wang_2019_CVPR}. Our proposed method called PromptMix, shown in Figure \ref{fig:overview}, is one such method that leverages the power of modern image synthesis tools, namely latent diffusion models, that are capable of generating realistic images given a text as input. PromptMix works by extracting text descriptions from existing datasets, using the extracted text as input to latent diffusion models to generate images that are similar to those in existing datasets, annotating the generated images using high-performing heavyweight networks, and mixing this \textit{fake dataset} with real datasets to improve the training of lightweight DNNs.

Through extensive experiments on crowd counting and monocular depth estimation tasks, five datasets and two lightweight network architectures, we showcase the effectiveness of PromptMix. Furthermore, we share the fake datasets created using PromptMix, which can be used by researchers to train other lightweight models. In addition, we introduce a novel ultra-lightweight architecture for crowd counting which can obtain a comparable performance to that of, or in some cases outperform, heavyweight networks when trained with PromptMix. Our code and datasets are publicly available.\footnote{\url{https://gitlab.au.dk/maleci/promptmix}}

The rest of this paper is organized as follows. Section \ref{sec:related} provides background regarding data boosting techniques in deep learning, information about the two computer vision tasks used in our experiments, as well as image captioning and synthesis methods which are utilized within PromptMix. Section \ref{sec:method} describes the proposed method in detail. Section \ref{sec:experiments} reports the results of our experiments and ablation studies. Finally, Section \ref{sec:conclusion} concludes the paper and discusses future research directions.

\section{Related Work}
\label{sec:related}

In this section, we first introduce crowd counting and monocular depth estimation, which are the tasks used in our experiments to evaluate the effectiveness of our proposed method. Subsequently, we recount existing approaches in the literature for artificially boosting the size of datasets for training. Finally, we provide a description of image captioning and image synthesis, which are used within the proposed method.

\subsection{Crowd Counting and Monocular Depth Estimation}
\label{sec:related_crowd_counting}

The goal of crowd counting \cite{gao2020cnn} is to count the total number of people present in a given image. The output of a crowd counting model is often in the form of a density map, which shows the density of the crowd for every pixel of the image. This density map can be integrated (summed) in order to obtain a total count for the scene. The labels in crowd counting datasets are often available in the form of head annotations, which specify the coordinates of the center of each person's head in the image. Head annotations can be converted into ground truth density maps by convolution with a gaussian kernel. Since each image may contain hundreds or thousands of people, manual annotation of each image can take a very long time. Therefore, crowd counting datasets are small and typically contain around one thousand images. Mean absolute error (MAE) and mean squared error (MSE) are two common metrics for evaluating crowd counting methods. MAE is used as a measure of performance, whereas MSE shows the robustness of the crowd counting model \cite{10.1145/3460426.3463628}.

Given a single image, monocular depth estimation (MDE) aims to find the distance relative to the camera for every pixel, resulting in a depth map \cite{Zhao2020}. Ground truth depth values are measured using a specialized hardware, such as LiDAR. However, the output of these devices often contains missing values due to shadows and reflections, which need to be interpolated in a post-processing step. Multiple evaluation metrics are commonly used for evaluating MDE methods in the literature, including root mean squared error (RMSE) and threshold accuracy $\delta_j$ defined as the ratio of predicted depth values $ \hat{y}_i $ that satisfy $ \max(\frac{y_i}{\hat{y}_i}, \frac{\hat{y}_i}{y_i}) < 1.25^j $, where $ y_i $ denotes the ground truth depth value at location $ i $ in the image. Three different threshold accuracies are commonly calculated based on $ j = \{1, 2, 3\} $ \cite{rudolph2022lightweight}. Higher $ j $ values result in more lenient metrics.

\subsection{Data Boosting Techniques in Deep Learning}
\label{sec:data_techniques}

The goal of data boosting techniques is to mitigate domain bias and overfitting issues that occur during the training of deep learning models, which often happen due to the small number of training examples. \textit{Data augmentation} \cite{Shorten2019} is a common technique where existing training examples are slightly modified each time they are sampled in a batch during training. For images, these modifications are typically simple operations such as rotation with a random angle, grayscaling color images, and horizontal flip that results in a mirror image. Even though the simplicity of these operations makes it easy to utilize data augmentation during training, it also leads to images that are not significantly different from the original ones, which limits the effectiveness of this approach.

DAGAN \cite{antoniou2017data} is a data augmentation technique that, 
instead of simple operations, uses generative adversarial networks (GANs) \cite{goodfellow2014generative} to generate images that are similar to the original image in the latent space. A disadvantage of this technique is that GANs are notoriously difficult to train \cite{Rombach_2022_CVPR}. Furthermore, DAGAN uses the same label as the original image for the generated image. This works well for the image classification task, since slight variations in the latent space are unlikely to produce images that belong to other classes. However, for tasks such as crowd counting, it might change the details of the scene such as the number and location of people in an unpredictable manner, making it impossible to use the same label. Techniques based on neural style transfer \cite{jackson2019style} circumvent this issue by only modifying the textures. However, since no other scene details such as the number of people or the perspective change, the benefits of style transfer is limited. Additionally, most style transfer methods use different painting styles for style transfer, which leads to non-photorealistic images.

The use of simulation-based techniques is an alternatives to data augmentation which aims to generate photorealistic synthetic data, which can be used for pre-training or be mixed with real data during training. Since the simulation is performed programmatically, obtaining labels is a straightforward procedure. Game engines, such as Unreal Engine\footnote{\url{https://www.unrealengine.com}}, have been used for this purpose \cite{8639113}. However, creating photorealistic scenes in game engines requires a great deal of manual work, including the creation of 3D meshes and textures for all objects in the scene, placement of the objects in relation to each other in a realistic way, and lighting configuration. Photorealistic video games such as Grand Theft Auto V (GTA V) can be used to reduce the amount of manual work \cite{Wang_2019_CVPR}, however, the downside is that video games might have undesired technical limitations. For instance, in GTA V, the number of people in a scene must be less than 256, and only 265 variations of person models exist. Moreover, even though GTA V can be used for crowd counting, a suitable photorealistic game may not exist for other computer vision tasks.

Another technique called \textit{Copycat} \cite{8489592} generates annotations for images from a completely different domain using an annotator deep neural network (DNN) trained on the target dataset. The resulting dataset is then mixed with the target dataset during training. The issue with Copycat is that it is often difficult to find images from a different domain that match properties of the target domain. For instance, images from an image classification or an object detection dataset cannot be used for crowd counting since they include very few people, whereas almost all images in crowd counting datasets contain hundreds or thousands of people. In addition, Copycat is very sensitive to the choice of dataset, annotator network and hyper-parameter values \cite{BAKHTIARNIA2022461}.

\subsection{Image Captioning and Image Synthesis}

Given an input image, the goal of image captioning is to obtain an informative text describing the contents of the image.
In this work, we use ClipCap \cite{mokady2021clipcap} in order to obtain captions from images in existing datasets. ClipCap utilizes CLIP, a pre-trained shared latent space for both image and text \cite{pmlr-v139-radford21a}, as well as GPT-2, a powerful language model \cite{radford2019language}; and a lightweight Transformer to simplify the training process and achieve state-of-the-art results.

Diffusion models (DMs) are the state-of-the-art for image synthesis at the time of this writing. Compared to GANs, DMs are easier to train and produce superior outputs. To train DMs, Gaussian noise is added to real images over several iterations until the output is pure noise. DMs then learn to reverse this process by denoising images step by step. This denoising process can be conditioned on text encodings in order to produce images from given texts, which are referred to as \textit{prompts}. Latent diffusion models (LDMs) significantly reduce the memory and computation requirements of DMs by performing the diffusion process in a lower dimensional latent space instead of the pixel space, and use an autoencoder to convert the generated output back to pixel space \cite{Rombach_2022_CVPR}. In this paper, we use a pre-trained open source implementation of LDMs called Stable Diffusion\footnote{\url{https://github.com/CompVis/stable-diffusion}} to generate images from obtained captions.

\section{Enhancing Lightweight Networks with PromptMix}
\label{sec:method}

We assume that for a given computer vision task, a target real dataset and at least one high-performing pre-trained heavyweight network exist. Note that the heavyweight network does not necessarily need to be trained on the target dataset, just trained on any dataset for the task at hand. The goal is to train a lightweight network for the task. The baseline approach is to simply train the lightweight architecture using the target dataset. PromptMix is an alternative training method that artificially boosts the size of the training set.

Figure \ref{fig:overview} shows an overview of the training process using PromptMix. PromptMix consists of six stages: prompt generation, prompt modification, image generation, image filtering, data mixing, and prompt filtering. The first stage extracts a text prompt $p^i_{\text{auto}}$ from each image $ I_i $ that exists in the training set of the target dataset using an image captioning DNN denoted by $ f_{\text{cap}} $, leading to the bag of automatically determined prompts $ P_{\text{auto}} = \{p^1_{\text{auto}}, \dots, p^N_{\text{auto}}\} $, where
$$
p^i_{\text{auto}} = f_{\text{cap}}(I_i), \thickspace 1 \leq i \leq N,
$$
and $ N $ is the total number of training examples. Text prompts may also come from dataset metadata. For instance, if it is known that the images are taken from interiors of several locations such as apartments, kitchens and bathrooms, prompts in the form of ``[location] interior'' can be manually added, resulting in the bag of manually determined prompts $ P_{\text{man}} $. The bag of candidate prompts $ P = \{p_1, \dots, p_M\} $ consists of all prompts in $ P_{\text{auto}} $ and $ P_{\text{man}} $.

\begin{figure}
\centering
\begin{subfigure}{.16\textwidth}
  \centering
  \includegraphics[width=.96\linewidth]{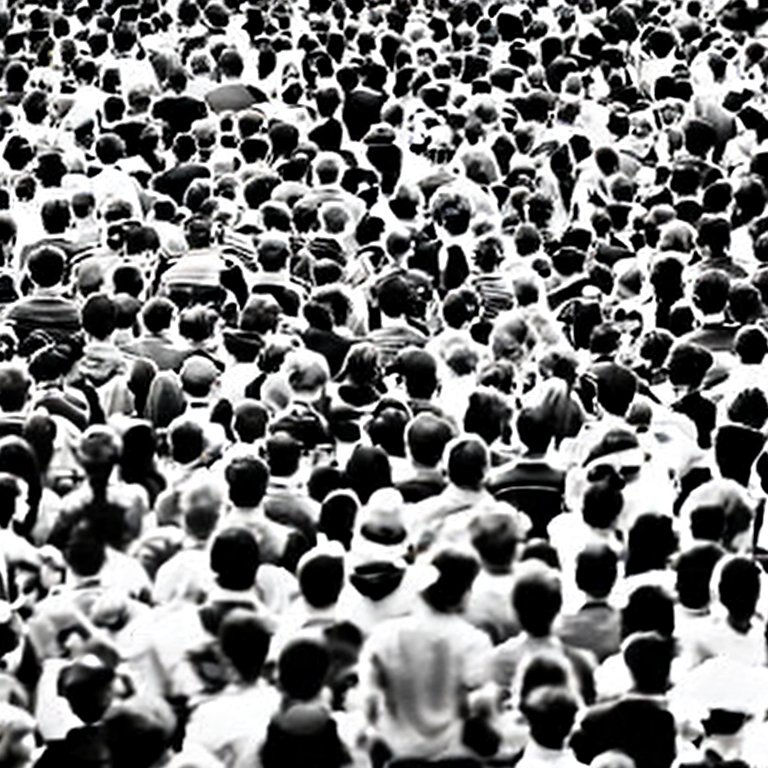}
\end{subfigure}%
\begin{subfigure}{.16\textwidth}
  \centering
  \includegraphics[width=.96\linewidth]{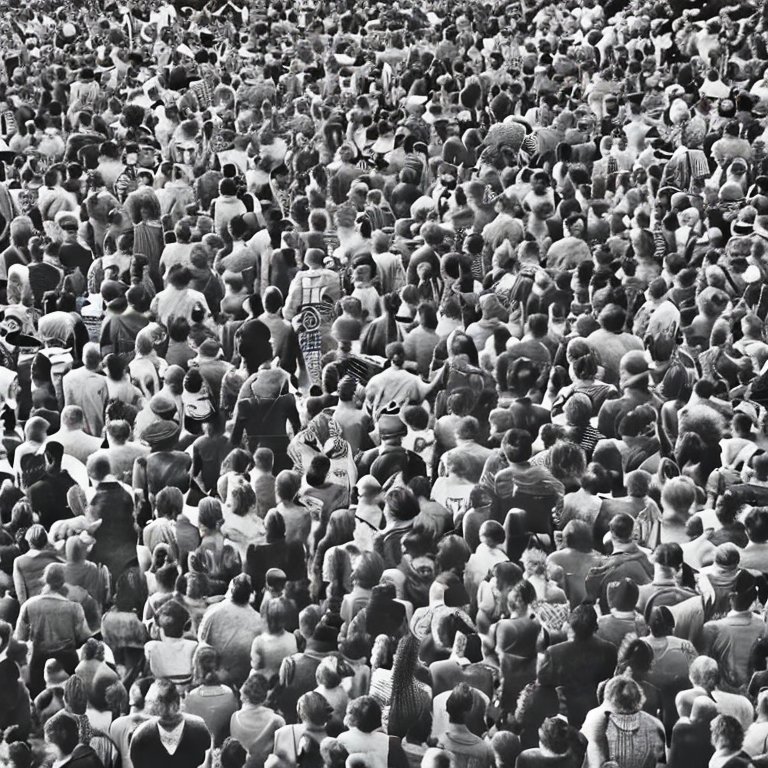}
\end{subfigure}
\begin{subfigure}{.16\textwidth}
  \centering
  \includegraphics[width=.96\linewidth]{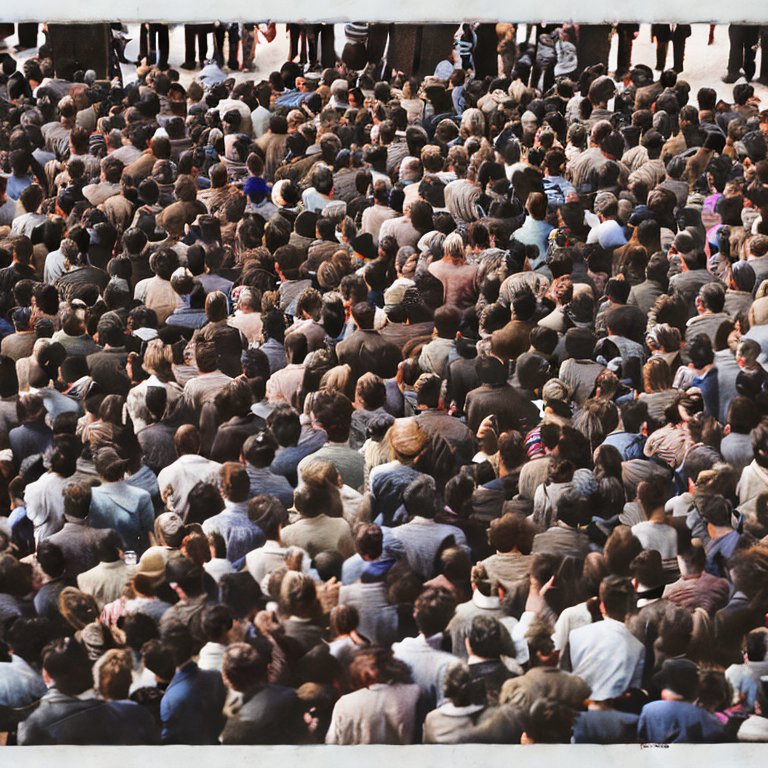}
\end{subfigure}
\caption{Left: example image generated with the prompt ``large crowd, black and white image''; Center: example image generated with the prompt ``large crowd, grayscale image''; Right: example image generated with the prompt ``large crowd, color image''.}
\label{fig:grayscale_vs_black_and_white}
\end{figure}

The second stage aims to improve the quality of candidate prompts. This can include manual inspection to rule out irrelevant prompts that may have been added to the bag due to erroneous outputs of image captioning DNN, although as we show in the experiments, this manual inspection is not necessary. More importantly, certain aspects of the image can be inferred automatically and emphasized in the prompt in the form of suffixes/prefixes. For instance, whether an image is grayscale or has color can be easily determined from the original image and reflected in the prompt as a suffix. We observed that using ``black and white image'' leads to near-monochrome images whereas ``grayscale image'' leads to images that utilize the full gray range, as shown in Figure \ref{fig:grayscale_vs_black_and_white}. $P' = \{p'_1, \dots, p'_{M'}\}, \thickspace M' \leq M$ is the bag of modified prompts which will be used in the next stage.

The third stage generates a fixed number $k$ of images $I^1_i, \dots I^k_i$ for each $ p'_i \in P'$ using a text-to-image diffusion model $f_{\text{dm}}$ where each image is generated using a different random seed. This operation can be summarized as
$$
I^g_i = f_{\text{dm}}(p'_i, r^g_i), \thickspace p'_i \in P', \thickspace g = \{1, \dots, k\},
$$
where each $r^g_i$ is a distinct random seed. Note that the width and height of the generated image should match that of the corresponding image in the target dataset. It is important that the image captioning DNN and the diffusion model use similar text encodings so that the generated images resemble real ones. Both ClipCap and Stable Diffusion use CLIP text encodings, however, with different configurations. Although ClipCap uses CLIP ViT-B/32 and Stable Diffusion uses CLIP ViT-L/14, these text encodings are very similar. This can be verified by reversing the order of captioning and generation operations, that is, selecting a text prompt $p'_i$, generating an image $I^g_i$ by using $p'_i$ as input for Stable Diffusion, then obtaining caption $\hat{p}'_i$ for $I^g_i$ using ClipCap, and comparing the edit distance of $p'_i$ and $\hat{p}'_i$. Testing various sample prompts of around ten words always resulted in a word edit distance of maximum two. All generated images are then annotated using all available pre-trained heavyweight networks. Thus, if $l$ annotator DNNs $f^1_{\text{ann}} , \dots, f^l_{\text{ann}}$ are available, for each generated image $I^g_i$, several annotations $A^g_{i,1} \dots A^g_{i,l}$ are obtained, where
$$
A^g_{i,j} = f^j_{\text{ann}}(I^g_i), \thickspace 1 \leq j \leq l.
$$

The fourth stage filters poor quality images based on the annotations. The intuition behind this approach is that since all annotator DNNs are high-performing models, the more they disagree on the predictions for a specific image, the more likely it is that the image falls outside the target domain. While this statement might not be true for individual images, as we show in the experiments, this method of filtering can effectively lead to less noise and overall better training. Formally, if $ l > 1 $, a quality score $ q^g_i $ is assigned to each image $ I^g_i $ based on the maximum relative difference between any two pairs of annotations, that is
$$
q^g_i = \frac{\max_{j \neq j}d(A^g_{i,j}, A^g_{i,j'})}{\max_jd(A^g_{i,j}, 0)},
$$
where $ d $ is a distance measure, for instance, the difference between the total count in crowd counting, defined as
$$
d(A, B) = |\sum_{i, j}A_{ij} - \sum_{i, j}B_{ij}|.
$$
The images are then sorted based on this quality score, and images inside the lowest $r_{\text{img}}$ percent are kept, while other images are discarded. This leads to the set of filtered images $\{I_1, \dots, I_{M''}\}, \thickspace M'' \leq kM' $ with annotations $ \{A^i_1, \dots, A^i_l\}$ corresponding to image $I_i$.

The fifth stage is to train the lightweight network with the help of generated images and annotations. To do this, we need to settle on only one annotation per image. As we show in the experiments, most choices will lead to improvements over the baseline. Annotations can even be aggregated, for instance, averaged. However, we found that the best configuration was to simply use the annotation from a heavyweight DNN trained on the target dataset. This leads to a \textit{fake dataset} $ D_{\text{fake}} = \{(I_i, A^i_j)|1 \leq i \leq M'' \} $ where $ j $ specifies the selected annotator heavyweight DNN. Since this fake dataset may contain several orders of magnitude more examples than the real dataset, using the entire fake dataset during training might introduce too much noise and hinder training. Instead,  a subset of training examples are randomly sampled from the fake dataset at each epoch $ t $, resulting in a sampled fake dataset $ D^t_{\text{sample}} $ where $ r_{\text{mix}} = |D^t_{\text{sample}}| $ is a hyper-parameter of the method. The training set used at epoch $ t $ is then $ D^t_{\text{epoch}} = D_{\text{real}} \cup D^t_{\text{sample}} $.

As shown in the ablation studies, some prompts may generate images that do not belong to the intended domain and, thus, impede the training. This can be the case especially for some manual prompts which are not extracted from the real dataset in a data-driven manner. To filter out such prompts, the optional last stage in PromptMix repeats the fifth stage for all prompts $ p_i $ but only using the images generated from that specific prompt, to obtain a validation error $ \text{MAE}^{p_i}_{\text{val}} $ for the lightweight model. Keep in mind that the validation error should be calculated only on a separate validation set which is used for model selection, and not the test set. Let $ \text{MAE}^{\text{base}}_{\text{val}} $ be the validation error of the lightweight model when it is trained only using the real dataset. If $ \text{MAE}^{p_i}_{\text{val}} > \text{MAE}^{\text{base}}_{\text{val}} $, prompt $ p_i $ and all its corresponding generated images are discarded. After this process is repeated for all prompts, the fifth stage is performed with the remaining fake dataset. We only recommend using prompt filtering for manual prompts, since it is a very time consuming operation.

The drawback of PromptMix is that it has many hyper-parameters and choices, namely whether to extract prompts manually, automatically or both, number of images generated per prompt ($ k $), heavyweight annotator DNNs ($ l $), image filtering ratio ($ r_{\text{img}} $), number of fake examples sampled at each epoch ($ r_{\text{mix}} $), and whether to use prompt filtering. However, as we show in the ablation studies, most setups offer improvements over the baseline. Furthermore, the power of PromptMix is that these hyper-parameters and choices allow for tuning the amount of manual labor, that is, the process could be completely automated end-to-end without any manual intervention, every single prompt and generated images can be manually curated and inspected, or anything in between.

PromptMix has several advantages over the techniques mentioned in Section \ref{sec:data_techniques}: it provides much more variation than data augmentation and neural style transfer, while avoiding annotation issues that arise with non-image classification tasks in DAGAN; unlike simulation-based methods, it can perform with little to no manual intervention; and as opposed to Copycat, it does not require an existing similar dataset and is not overly sensitive to hyper-parameters. Additionally, prompts are human-readable and thus much easier to interpret and control compared to the latent space in GAN-based data augmentation methods.

\section{Experiments}
\label{sec:experiments}

\subsection{Setup}

\subsubsection{Crowd Counting}

We use four datasets to evaluate PromptMix on crowd counting. The Shanghai Tech Part A (SHTA) dataset \cite{Zhang_2016_CVPR} consists of 300 training and 182 test images with an average resolution of 868$\times$589 pixels, scraped from the web. The Shanghai Tech Part B (SHTB) dataset \cite{Zhang_2016_CVPR} contains 400 training and 316 test images with a resolution of 1,024$\times$768 pixels, taken from real-world surveillance cameras. The UCF-QNRF dataset \cite{Idrees_2018_ECCV} includes 1,201 training and 334 test images with an average resolution of 2,902$\times$2,013 pixels, collected from the web. The PANDA-Crowd dataset \cite{Wang_2020_CVPR} consists of 45 gigapixel images with an average resolution of 26,382$\times$14,767 pixels taken from three real-world scenes: an airport, a marathon and a graduation ceremony. SHTA, SHTB and UCF-QNRF are widely used in the crowd counting literature, and PANDA-Crowd is a new and challenging dataset due to the low number of images and their extremely high resolution.

Since UCF-QNRF has very high-resolution images, similar to other methods in the literature, we resize its images to fit within a 1,920$\times$1,920 box while preserving the original aspect ratio. We resize PANDA-Crowd images to 2,560$\times$1,440, which is the maximum size that fits our GPU memory. We randomly select 20\%, 10\% and 10\% of the training set of SHTA, SHTB and UCF-QNRF, respectively, to use as validation set. PANDA-Crowd does not have pre-determined training and test sets, therefore, we take 30 images for training, 6 for validation and 9 for test. PANDA-Crowd labels are bounding boxes instead of head annotations, thus we match the Gaussian kernel size with the bounding box when generating ground truth density maps. The size used for the Gaussian kernel for other datasets is 15, and $\sigma = 4$.

As image filtering is not used in our main experiments, we only use a single heavyweight annotator DNN trained on the target dataset for each experiment. For SHTA, SHTB and PANDA experiments, we use the SASNet architecture \cite{Song_Wang_Wang_Tai_Wang_Li_Wu_Ma_2021} which uses the first ten layers of VGG-16 \cite{DBLP:journals/corr/SimonyanZ14a} as a feature extractor and combines these features across multiple scales. We use pre-trained weights provided by the authors for SHTA and SHTB. There are no pre-trained weights available for PANDA-Crowd, therefore, we fine-tune SASNet on PANDA-Crowd to use for annotation. SASNet does not provide pre-trained weights for UCF-QNRF, and we were not able to replicate the performance reported in their paper by fine-tuning. Therefore, we use DM-Count \cite{NEURIPS2020_118bd558} with a VGG-19 backbone as heavyweight annotator DNN for UCF-QNRF experiments.

For the lightweight DNN, we conduct experiments with two different architectures. The first one is CSRNet \cite{Li_2018_CVPR}
which also uses the first ten layers of VGG-16 \cite{DBLP:journals/corr/SimonyanZ14a} pre-trained on ImageNet \cite{5206848} as feature extractor. Compared to more modern crowd counting architectures such as SASNet, CSRNet requires much less computation and memory as it only has six dilated convolution layers besides the feature extractor. However, it is still challenging to run CSRNet on some computationally-restricted IoT devices. Even though VGG is an excellent feature extractor for crowd counting \cite{gao2020cnn}, it adds a lot of overhead to the architecture and is the main reason why most modern crowd counting are not lightweight. However, it has been recently shown that shallow ResNets \cite{He_2016_CVPR} are also good feature extractors \cite{kumar2022do}. Based on this fact, by replacing the feature extractor in CSRNet with the first 9 layers of ResNet18 pretrained on ImageNet, and restricting the number of channels in the dilated convolution layers to 128 down from 512, we introduce an ultra-lightweight version of CSRNet called ResCSRNet, which is also used in our experiments.

In all crowd counting experiments, prompts are automatically extracted from the target datasets and no manual prompts are used. Prompts are modified by manually adding the prefix ``large crowd'' to ensure the images contain dense crowds, and by automatically adding a suffix based on whether the original image was grayscale or color. We generate 10 images for each prompt in SHTA and STHB and PANDA-Crowd experiments (that is, $ k = 10 $). However, due to the large number of images in UCF-QNRF as well as their high resolution, our GPU is only capable of generating one image for each prompt within reasonable time. We set $ r_{\text{mix}} $ to 36 for PANDA-Crowd and 500 for all other datasets. It is important to note that JPEG compression can have a drastic effect on the performance of crowd counting \cite{9921939, bakhtiarnia2022crowd}, and most images provided in crowd counting datasets are compressed using JPEG. Therefore, our generated data needs to be compressed with the same quality factor (QF) as the original dataset. The QF of SHTA, SHTB and UCF-CC-50 is 75 for all images. The QF of UCF-QNRF varies from image to image, therefore, we use the average value which is 91. Since the main objective of the proposed method is to improve the performance, we use the MAE metric to evaluate crowd counting models.

\subsubsection{Monocular Depth Estimation}

We use the NYU Depth V2 \cite{Silberman:ECCV12}, a widely used monocular depth estimation dataset, which contains 1,449 images with a resolution of 640$ \times $480 pixels. Since NYU Depth V2 does not specify pre-determined training and test sets, we use 70\% for training, 10\% for validation and 20\% for test. We use AdaBins \cite{Bhat_2021_CVPR} pre-trained on NYU Depth V2 as the heavyweight annotator DNN, and GuideDepth \cite{rudolph2022lightweight} as the lightweight network.

The metadata for NYU Depth V2 includes precise metadata about where the images were taken. Thus we use three manual prompts, namely ``apartment interior'', ``bedroom interior'' and ``bathroom'' interior, and generate 2,500 images per prompt, ($ k = 2500 $). Using image captioning resulted in repetitive captions similar to these prompts. We set $ r_{\text{mix}} $ to 1000 and use the $ \delta_1 $, $ \delta_2 $, $ \delta_3 $ as well as RMSE metrics to evaluate monocular depth estimation models.

All crowd counting and monocular depth estimation models were trained using the AdamW \cite{loshchilov2017decoupled} optimizer with a weight decay of $ 10^{-4} $ for 500 epochs. Hyperparameter optimization was done for all experiments to find the best learning rate among $ \{10^{-3}, 10^{-4}, 10^{-5}, 10^{-6}\} $. The learning rate was multiplied by a factor of $ 0.99 $ every epoch. The random seed used for selecting validation set is 42 for all datasets, and the selection codes are available in our code repository. All experiments were conducted on Nvidia A6000 GPUs using the PyTorch library \cite{NEURIPS2019_bdbca288}. The batch size for SHTB and NYU Depth V2 experiments was 32. A batch size of 1 was used for SHTA and UCF-QNRF as they have variable image sizes and PyTorch is not capable of handling batches containing different sizes. The batch size of experiments with the PANDA-Crowd was also set to 1 due to GPU memory limitations.

\subsection{Results}

The results of crowd counting and monocular depth estimation are summarized in Tables \ref{tab:cc_results} and \ref{tab:mde_results}, respectively. In all experiments shown in these Tables, when the lightweight network is trained using PromptMix, it obtains a performance that is significantly better than that of the baseline. Note that the resolution of PANDA-Crowd is too high for the ultra-lightweight ResCSRNet, therefore, it is not able to converge and produce reasonable density maps after training. As a result, Table \ref{tab:cc_results} does not include experiments with ResCSRNet on the PANDA-Crowd dataset.

\begin{table}[htbp]
\caption{Performance of PromptMix for the crowd counting task on the Shanghai Tech Part A, Shanghai Tech Part B, UCF-QNRF and PANDA datasets. The lowest error for each experiment is highlighted.}
\begin{center}
\resizebox{\linewidth}{!}{
\begin{tabular}{ l c c c c } 
\hline
Method & SHTB & SHTA & UCF-QNRF & PANDA-Crowd\\
& MAE$\downarrow$ & MAE$\downarrow$ & MAE$\downarrow$ & MAE$\downarrow$\\
\hline
\hline
ResCSRNet (baseline) & 11.63 & 124.41 & 196.93 & -\\
ResCSRNet+PromptMix (ours) & \textbf{8.57} & \textbf{100.75} & \textbf{187.93} & -\\
\hline
CSRNet (baseline)  & 10.60 & 68.2 & 122.55 & 137.72\\
CSRNet+PromptMix (ours) & \textbf{8.68} & \textbf{65.93} & \textbf{105.91} & \textbf{110.34}\\
\hline
\end{tabular}
}
\end{center}
\label{tab:cc_results}
\end{table}

\begin{table}[htbp]
\caption{Performance of PromptMix for the monocular depth estimation task on the NYU Depth V2 dataset. The best performance for each experiment is highlighted. Note that for the $\delta_1, \dots \delta_3$ metrics, larger numbers are more accurate, whereas for the RMS metric, smaller numbers are better.}
\begin{center}
\begin{tabular}{ l c c c c } 
\hline
Method & $\delta_1 \uparrow$ & $\delta_2 \uparrow$ & $\delta_3 \uparrow$ & RMSE$\downarrow$\\ 
\hline
\hline
GuideDepth (baseline) & 0.754 & 0.943 & 0.987 & 0.639\\
GuideDepth+PromptMix (ours) & \textbf{0.780} & \textbf{0.954} & \textbf{0.989} & \textbf{0.601}\\
\hline
\end{tabular}
\end{center}
\label{tab:mde_results}
\end{table}

\begin{figure}[!t]
\begin{center}
\begin{tabular}{@{} c c @{}}
\includegraphics[width=.23\textwidth]{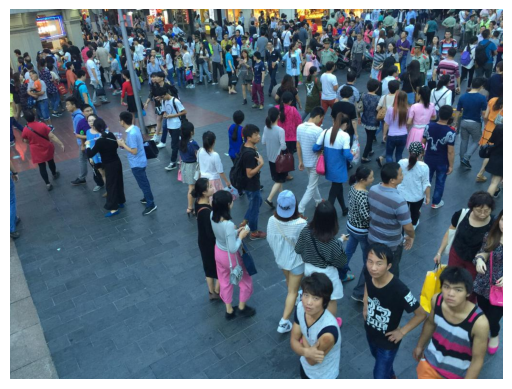}&
\includegraphics[width=.23\textwidth]{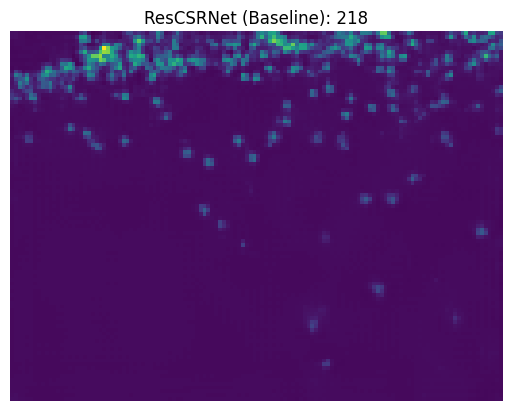}\\
(a) & (b)\\
\includegraphics[width=.23\textwidth]{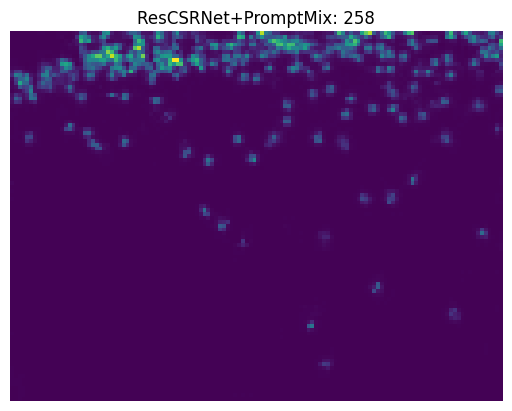}&
\includegraphics[width=.23\textwidth]{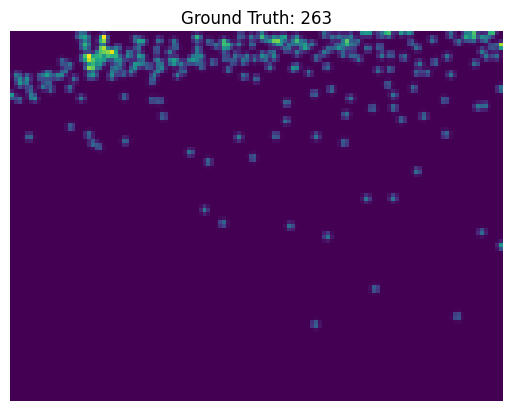}\\
(c) & (d)\\
\end{tabular}
\end{center}
\caption{In the crowd counting task, the network trained with PromptMix leads to a sharper density map that is much closer to the ground truth compared to the baseline: (a) example image from the Shanghai Tech Part B dataset, (b) density map and count from baseline, (c) density map and count of PromptMix, and (d) ground truth density map and count.}
\label{fig:cc_sample}
\end{figure}

\begin{figure}
\begin{center}
\begin{tabular}{@{} c c @{}}
\includegraphics[width=.23\textwidth]{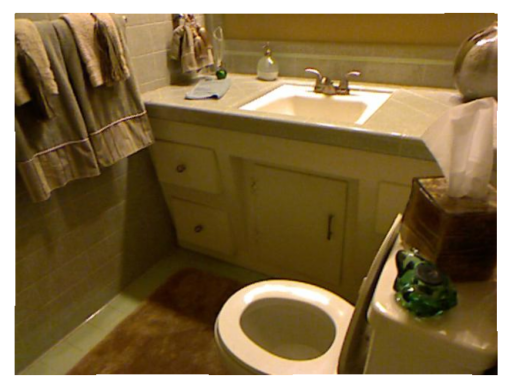}&
\includegraphics[width=.23\textwidth]{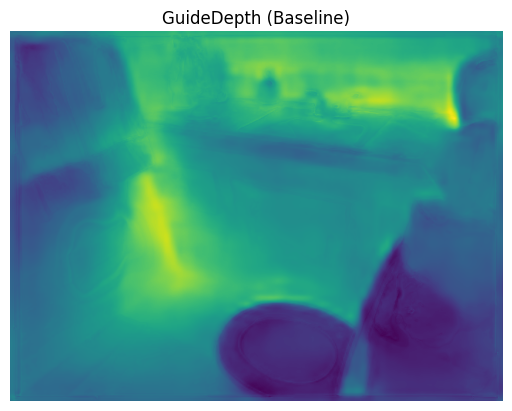}\\
(a) & (b)\\
\includegraphics[width=.23\textwidth]{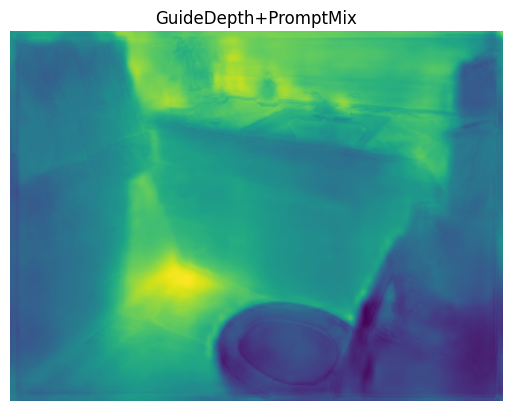}&
\includegraphics[width=.23\textwidth]{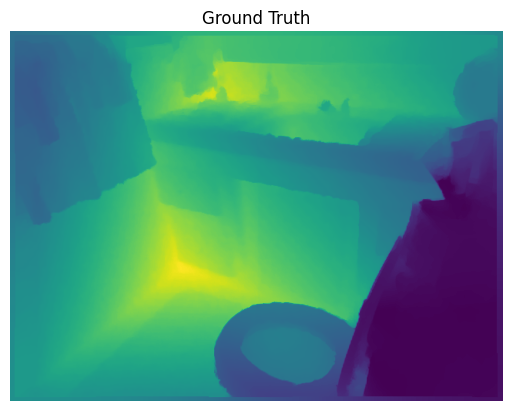}\\
(c) & (d)\\
\end{tabular}
\end{center}
\caption{In the monocular depth estimation task, the network trained with PromptMix can pick up more intricate details from the scene compared to the baseline, such as the toilet rim and sink outline: (a) example image from the NYU Depth V2 dataset, (b) output from baseline, (c) output of PromptMix, and (d) ground truth depth map.}
\label{fig:mde_sample}
\end{figure}

Figures \ref{fig:cc_sample} and \ref{fig:mde_sample} depict an example input image as well as the inference results of the baseline and PromptMix compared to the ground truth, for crowd counting and monocular depth estimation, respectively. In both Figures, it can be observed that the lightweight network trained with PromptMix is able to produce a sharper density or depth map.
\begin{table}[!t]
\caption{Efficiency vs. performance of different methods for the crowd counting task on the Shanghai Tech Part B dataset.}
\begin{center}
\resizebox{\linewidth}{!}{
\begin{tabular}{ l c c c c } 
\hline
Method & GMac$\downarrow$ for 1,024$\times$768 input & Params$\downarrow$ (M) & MAE$\downarrow$\\ 
\hline
\hline
ResCSRNet+PromptMix & 26.54 & 1.57 & 8.57\\
CSRNet & 325.34 & 16.26 & 10.60\\
SASNet & 698.72 & 38.90 & 6.35\\
\hline
\end{tabular}
}
\end{center}
\label{tab:efficiency_vs_accuracy}
\end{table}

Table \ref{tab:efficiency_vs_accuracy} shows the performance-efficiency tradeoff for ResCSRNet, CSRNet and SASNet. Despite the fact that ResCSRNet uses 12.26$\times$ less computation and has 10.36$\times$ lower number of parameters than CSRNet, when trained with PromptMix, ResCSRNet outperforms the original CSRNet by 19.15\% on the SHTB dataset.

\subsection{Ablation Studies}\label{sec:ablation}
Table \ref{tab:ablation_numbers} explores the effect of the total number of generated images as well as the number of images sampled per epoch on the outcome of PromptMix. Comparing second and fourth rows shows that sampling too many generated images per epoch can introduce noise and impair performance. Comparing the last three rows, it is evident that having more generated images to sample from, which leads to more unique data at each epoch, provides benefits. Finally, comparing the baseline in the first row with all other rows, it can be observed that many different hyper-parameters can result in performance improvements, even if they are not perfectly tuned. 

\begin{table}[htbp]
\caption{Ablation study on number of generated and resampled images used during training. The ResCSRNet architecture is trained using PromptMix on the Shanghai Tech Part B dataset. The images are generated using only a single manual prompt ``large crowd''.}
\begin{center}
\begin{tabular}{ l l c } 
\hline
$k$ & $r_{\text{mix}}$ & MAE$\downarrow$\\ 
\hline
\hline
0 & 0 & 11.63\\
1,000 & 1,000 & 10.22\\
500 & 500 & 9.96\\
1,000 & 500 & 9.33\\
10,000 & 500 & 9.05\\
\hline
\end{tabular}
\end{center}
\label{tab:ablation_numbers}
\end{table}

As discussed in Section \ref{sec:method}, the heavyweight annotator network does necessarily not need to be trained on the target dataset in order to be effective, although pre-training on the target dataset tends to achieve the best performance. Table \ref{tab:ablation_pre_training} shows that regardless of the choice of dataset used for pre-training, the error of PromptMix is lower than the baseline.

\begin{table}[htbp]
\caption{Ablation study on the dataset used for pre-training heavyweight network, and number of generated and resampled images used during training. The ResCSRNet architecture is trained using PromptMix on the Shanghai Tech Part B dataset. The architecture of the heavyweight network is SASNet. Images are generated using only a single manual prompt ``large crowd'', and 500 are resampled each epoch.}
\begin{center}
\begin{tabular}{ c c c } 
\hline
$k$ & Pre-Training Dataset & MAE$\downarrow$\\ 
\hline
\hline
0 & - & 11.63\\
\hline
1,000 & SHTA  & 10.33\\
1,000 & SHTB & 9.33\\
\hline
10,000 & SHTA & 10.11\\
10,000 & SHTB & 9.05\\
\hline
\end{tabular}
\end{center}
\label{tab:ablation_pre_training}
\end{table}

Text-to-image diffusion models currently lead to low quality generated faces, therefore, face restoration methods such as GFP-GAN \cite{Wang_2021_CVPR} are often used as a post-processing step to improve the quality of generated faces. Even though GFP-GAN is capable of slightly improving the quality of faces in crowds, using face restoration does not improve the performance of PromptMix based on the first two rows of Table \ref{tab:ablation_attributes}. Moreover, Stable Diffusion uses half-precision floating-points by default in order to reduce the memory and computation costs. However, based on the first and last rows of the table, using full-precision does not improve PromptMix.

\begin{table}[htbp]
\caption{Ablation study on attributes of the generated images. The ResCSRNet architecture is trained using PromptMix on the Shanghai Tech Part B dataset. 1,000 images are generated using only a single manual prompt ``large crowd'', and 500 are resampled each epoch.}
\begin{center}
\begin{tabular}{ l c c } 
\hline
Floating-Point Precision & Face Restoration & MAE$\downarrow$\\ 
\hline
\hline
Half-Precision (float16) & - & 9.33\\
Half-Precision (float16) & \checkmark & 10.13\\
Full-Precision (float32) & - & 10.40\\
\hline
\end{tabular}
\end{center}
\label{tab:ablation_attributes}
\end{table}

Manual prompts should be carefully selected to match the targeted domain. For instance, images of train stations and Mecca are common in crowd counting datasets, however, SHTB does not include any images of Mecca. The first two rows of Table \ref{tab:ablation_manual} show that if a manual prompt does not match the target domain, PromptMix can even harm the performance. However, prompt and image filtering can help in scenarios where irrelevant manual prompts may exist. Comparing the third and fourth rows in the Table, it can be observed that using prompt filtering boosts performance even if some prompts do not match the domain. Comparing the third and last rows in the Table shows that using image filtering can provide a similar benefit. In addition, the last three rows show that moderate image filtering is a better choice compared to aggressively filtering out most images.

\begin{table}[htbp]
\caption{Ablation study on manual prompts. The ResCSRNet architecture is trained using PromptMix on the Shanghai Tech Part B dataset. 1,000 images are generated per prompt, and 500 are resampled each epoch. Prompt filtering is done using two SASNet models trained on Shanghai Tech Parts A and B. The ten locations chosen based on popular crowd counting datasets are: ``a public speech'', ``Mecca'', ``a bus station'', ``a train station'', ``a ceremony'', ``a concert'', ``a marathon'', ``an airport'', ``a stadium'', and ``the streets''.}
\begin{center}
\resizebox{\linewidth}{!}{
\begin{tabular}{ c l c c c } 
\hline
$|P_{\text{man}}|$ & Prompt Format & Prompt Filtering & $r_{\text{img}}$ & MAE$\downarrow$\\ 
\hline
\hline
1 & ``large crowd in a train station'' & - & 100\% & 8.88\\
1 & ``large crowd in mecca'' & - & 100\% & 12.08\\ 
10 & ``large crowd in [location]'' & - & 100\% & 10.40\\
10 & ``large crowd in [location]'' & \checkmark & 100\% & 8.64\\
10 & ``large crowd in [location]'' & - & 10\% & 10.49\\
10 & ``large crowd in [location]'' & - & 40\% & 9.54\\
10 & ``large crowd in [location]'' & - & 60\% & 8.75\\
\hline
\end{tabular}
}
\end{center}
\label{tab:ablation_manual}
\end{table}

\section{Conclusion}
\label{sec:conclusion}

We presented PromptMix, a novel approach which utilizes image synthesis to boost deep learning datasets. Through extensive experiments on two tasks, five datasets, and two lightweight architectures, we showed that PromptMix can significantly improve the performance of lightweight DNNs. Using text as an intermediate representation, PromptMix is able to provide an explainable mechanism to modify the process if the generated images do not match the targeted domain. Additionally, we introduced tools such as prompt modification, image filtering and prompt filtering as well as several best practices in order to fully leverage PromptMix.

Although we test PromptMix only on dense regression tasks in this paper, it is likely to benefit other computer vision tasks as well, since no part of PromptMix explicitly relies on dense regression. The reason PromptMix is useful for such tasks is due to the fact that acquiring labels is difficult in these cases, therefore the size of the datasets is usually small compared to other vision tasks such as image classification. PromptMix could potentially be useful in image classification if the number of generated images is scaled to tens of millions. However, this is beyond our available computational resources.

\bibliographystyle{IEEEtran}
\bibliography{references.bib}

\begin{thebibliography}{10}
\providecommand{\url}[1]{#1}
\csname url@samestyle\endcsname
\providecommand{\newblock}{\relax}
\providecommand{\bibinfo}[2]{#2}
\providecommand{\BIBentrySTDinterwordspacing}{\spaceskip=0pt\relax}
\providecommand{\BIBentryALTinterwordstretchfactor}{4}
\providecommand{\BIBentryALTinterwordspacing}{\spaceskip=\fontdimen2\font plus
\BIBentryALTinterwordstretchfactor\fontdimen3\font minus
  \fontdimen4\font\relax}
\providecommand{\BIBforeignlanguage}[2]{{%
\expandafter\ifx\csname l@#1\endcsname\relax
\typeout{** WARNING: IEEEtran.bst: No hyphenation pattern has been}%
\typeout{** loaded for the language `#1'. Using the pattern for}%
\typeout{** the default language instead.}%
\else
\language=\csname l@#1\endcsname
\fi
#2}}
\providecommand{\BIBdecl}{\relax}
\BIBdecl

\bibitem{https://doi.org/10.1002/itl2.187}
S.~Bhattacharya, S.~R.~K. Somayaji \emph{et~al.}, ``A review on deep learning
  for future smart cities,'' \emph{Internet Technology Letters}, 2022.

\bibitem{DBLP:journals/corr/HowardZCKWWAA17}
A.~G. Howard, M.~Zhu \emph{et~al.}, ``Mobilenets: Efficient convolutional
  neural networks for mobile vision applications,'' \emph{CoRR}, 2017.

\bibitem{8578572}
M.~Sandler, A.~Howard \emph{et~al.}, ``Mobilenetv2: Inverted residuals and
  linear bottlenecks,'' \emph{CVPR}, 2018.

\bibitem{9008835}
A.~Howard, M.~Sandler \emph{et~al.}, ``Searching for mobilenetv3,''
  \emph{CVPR}, 2019.

\bibitem{LeCun2015}
Y.~LeCun, Y.~Bengio, and G.~Hinton, ``Deep learning,'' \emph{Nature}, 2015.

\bibitem{antoniou2017data}
A.~Antoniou, A.~Storkey, and H.~Edwards, ``Data augmentation generative
  adversarial networks,'' \emph{arXiv}, 2017.

\bibitem{Wang_2019_CVPR}
Q.~Wang, J.~Gao \emph{et~al.}, ``Learning from synthetic data for crowd
  counting in the wild,'' \emph{CVPR}, 2019.

\bibitem{gao2020cnn}
G.~Gao, J.~Gao \emph{et~al.}, ``Cnn-based density estimation and crowd
  counting: A survey,'' \emph{arXiv}, 2020.

\bibitem{10.1145/3460426.3463628}
F.~Dai, H.~Liu \emph{et~al.}, ``Dense scale network for crowd counting,''
  \emph{ICMR}, 2021.

\bibitem{Zhao2020}
C.~Zhao, Q.~Sun \emph{et~al.}, ``Monocular depth estimation based on deep
  learning: An overview,'' \emph{Science China Technological Sciences}, 2020.

\bibitem{rudolph2022lightweight}
M.~Rudolph, Y.~Dawoud \emph{et~al.}, ``Lightweight monocular depth estimation
  through guided decoding,'' \emph{ICRA}, 2022.

\bibitem{Shorten2019}
C.~Shorten and T.~M. Khoshgoftaar, ``A survey on image data augmentation for
  deep learning,'' \emph{Journal of Big Data}, 2019.

\bibitem{goodfellow2014generative}
I.~Goodfellow, J.~Pouget-Abadie \emph{et~al.}, ``Generative adversarial nets,''
  \emph{NeurIPS}, 2014.

\bibitem{Rombach_2022_CVPR}
R.~Rombach, A.~Blattmann \emph{et~al.}, ``High-resolution image synthesis with
  latent diffusion models,'' \emph{CVPR}, 2022.

\bibitem{jackson2019style}
P.~T. Jackson, A.~Atapour-Abarghouei \emph{et~al.}, ``Style augmentation: Data
  augmentation via style randomization,'' \emph{CVPR Workshops}, 2019.

\bibitem{8639113}
G.~Schröder, T.~Senst \emph{et~al.}, ``Optical flow dataset and benchmark for
  visual crowd analysis,'' \emph{AVSS}, 2018.

\bibitem{8489592}
J.~R. Correia-Silva, R.~F. Berriel \emph{et~al.}, ``Copycat cnn: Stealing
  knowledge by persuading confession with random non-labeled data,''
  \emph{IJCNN}, 2018.

\bibitem{BAKHTIARNIA2022461}
A.~Bakhtiarnia, Q.~Zhang, and A.~Iosifidis, ``Single-layer vision transformers
  for more accurate early exits with less overhead,'' \emph{Neural Networks},
  2022.

\bibitem{mokady2021clipcap}
R.~Mokady, A.~Hertz, and A.~H. Bermano, ``Clipcap: Clip prefix for image
  captioning,'' \emph{arXiv}, 2021.

\bibitem{pmlr-v139-radford21a}
A.~Radford, J.~W. Kim \emph{et~al.}, ``Learning transferable visual models from
  natural language supervision,'' \emph{ICML}, 2021.

\bibitem{radford2019language}
A.~Radford, J.~Wu \emph{et~al.}, ``Language models are unsupervised multitask
  learners,'' \emph{OpenAI blog}, 2019.

\bibitem{Zhang_2016_CVPR}
Y.~Zhang, D.~Zhou \emph{et~al.}, ``Single-image crowd counting via multi-column
  convolutional neural network,'' \emph{CVPR}, 2016.

\bibitem{Idrees_2018_ECCV}
H.~Idrees, M.~Tayyab \emph{et~al.}, ``Composition loss for counting, density
  map estimation and localization in dense crowds,'' \emph{ECCV}, 2018.

\bibitem{Wang_2020_CVPR}
X.~Wang, X.~Zhang \emph{et~al.}, ``Panda: A gigapixel-level human-centric video
  dataset,'' \emph{CVPR}, 2020.

\bibitem{Song_Wang_Wang_Tai_Wang_Li_Wu_Ma_2021}
Q.~Song, C.~Wang \emph{et~al.}, ``To choose or to fuse? scale selection for
  crowd counting,'' \emph{AAAI}, 2021.

\bibitem{DBLP:journals/corr/SimonyanZ14a}
K.~Simonyan and A.~Zisserman, ``Very deep convolutional networks for
  large-scale image recognition,'' \emph{ICLR}, 2015.

\bibitem{NEURIPS2020_118bd558}
B.~Wang, H.~Liu \emph{et~al.}, ``Distribution matching for crowd counting,''
  \emph{NeurIPS}, 2020.

\bibitem{Li_2018_CVPR}
Y.~Li, X.~Zhang, and D.~Chen, ``Csrnet: Dilated convolutional neural networks
  for understanding the highly congested scenes,'' \emph{CVPR}, 2018.

\bibitem{5206848}
J.~Deng, W.~Dong \emph{et~al.}, ``Imagenet: A large-scale hierarchical image
  database,'' \emph{CVPR}, 2009.

\bibitem{He_2016_CVPR}
K.~He, X.~Zhang \emph{et~al.}, ``Deep residual learning for image
  recognition,'' \emph{CVPR}, 2016.

\bibitem{kumar2022do}
M.~Kumar, N.~Houlsby \emph{et~al.}, ``Do better imagenet classifiers assess
  perceptual similarity better?'' \emph{TMLR}, 2022.

\bibitem{9921939}
A.~Bakhtiarnia, B.~Leporowski \emph{et~al.}, ``Analysis of the effect of
  low-overhead lossy image compression on the performance of visual crowd
  counting for smart city applications,'' \emph{ISC2}, 2022.

\bibitem{bakhtiarnia2022crowd}
A.~Bakhtiarnia, Q.~Zhang, and A.~Iosifidis, ``Crowd counting on heavily
  compressed images with curriculum pre-training,'' \emph{arXiv}, 2022.

\bibitem{Silberman:ECCV12}
P.~K. Nathan~Silberman, Derek~Hoiem and R.~Fergus, ``Indoor segmentation and
  support inference from rgbd images,'' \emph{ECCV}, 2012.

\bibitem{Bhat_2021_CVPR}
S.~F. Bhat, I.~Alhashim, and P.~Wonka, ``Adabins: Depth estimation using
  adaptive bins,'' \emph{CVPR}, 2021.

\bibitem{loshchilov2017decoupled}
I.~Loshchilov and F.~Hutter, ``Decoupled weight decay regularization,''
  \emph{ICLR}, 2019.

\bibitem{NEURIPS2019_bdbca288}
A.~Paszke, S.~Gross \emph{et~al.}, ``Pytorch: An imperative style,
  high-performance deep learning library,'' \emph{NeurIPS}, 2019.

\bibitem{Wang_2021_CVPR}
X.~Wang, Y.~Li \emph{et~al.}, ``Towards real-world blind face restoration with
  generative facial prior,'' \emph{CVPR}, 2021.

\end{thebibliography}

\end{document}